# 4D Attention-based Neural Network for EEG Emotion Recognition

Guowen Xiao[1] • Mengwen Ye[2] • Bowen Xu[1] • Zhendi Chen[1] • Quansheng Ren[1*]

**Abstract**

Electroencephalograph (EEG) emotion recognition is a significant task in the brain-computer interface field. Although many deep learning methods are proposed recently, it is still challenging to make full use of the information contained in different domains of EEG signals. In this paper, we present a novel method, called four-dimensional attention-based neural network (4D-aNN) for EEG emotion recognition. First, raw EEG signals are transformed into 4D spatial-spectral-temporal representations. Then, the proposed 4D-aNN adopts spectral and spatial attention mechanisms to adaptively assign the weights of different brain regions and frequency bands, and a convolutional neural network (CNN) is utilized to deal with the spectral and spatial information of the 4D representations. Moreover, a temporal attention mechanism is integrated into a bidirectional Long Short-Term Memory (LSTM) to explore temporal dependencies of the 4D representations. Our model achieves state-of-the-art performance on the SEED dataset under intra-subject splitting. The experimental results have shown the effectiveness of the attention mechanisms in different domains for EEG emotion recognition.

[1]Department of Electronics,
Peking University, Beijing, China
[2]School of Electrical Engineering,
Beijing Jiaotong University, Beijing, China
*Corresponding author: Quansheng Ren (Email: qsren@pku.edu.cn)

# Introduction

Emotion plays an important role in daily life and is closely related to human behavior and cognition (Dolan 2002). As one of the most significant research topics of affective computing, emotion recognition has received increasing attention in recent years for its applications of disease detection (Bamdad et al. 2015; Figueiredo et al. 2019), human-computer interaction (Fiorinia et al. 2020; Katsigiannis and Ramzan 2017), and workload estimation (Blankertz et al. 2016). In general, emotion recognition methods can be divided into two categories (Mühl et al. 2014). One is based on external emotion responses including facial expressions and gestures(Yan et al. 2016), and the other is based on internal emotion responses including electroencephalograph (EEG) and electrocardiography (ECG) (Zheng et al. 2017). Neuroscientific researches have shown that some major brain cortex regions are closely related to emotions, making it possible to decode emotions based on EEG (Brittona et al. 2006; Lotfia and Akbarzadeh-T 2014). EEG is non-invasive, portable, and inexpensive so that it has been widely used in the field of brain-computer interfaces (BCIs) (Pfurtscheller et al. 2010). Besides, EEG signals contain various spatial, spectral, and temporal information about emotions evoked by specific stimulation patterns. Therefore, more and more researchers concentrate on EEG emotion recognition recently (Alhagry et al. 2017; Li and Lu 2009).

Traditional EEG emotion recognition methods usually extract hand-crafted features from EEG signals first and then adopt shallow models to classify the emotion features. EEG emotion features can be extracted from the time domain, frequency domain, and time-frequency domain. Jenke et al. conduct a comprehensive survey on EEG feature extraction methods by using machine learning techniques on a self-recorded dataset (Jenke et al. 2014). For classifying the extracted emotion features, many researchers have adopted machine learning methods over the past few years (Kim et al. 2013). Li et al. apply a linear support vector machine (SVM) to classify emotion features extracted from the gamma frequency band (Li and Lu 2009). Duan et al. extract differential entropy (DE) features, which are superior to representing emotion states in EEG signals (Shi et al. 2013), from multichannel EEG data and combine a k-Nearest Neighbor (KNN) with SVM to classify the DE features (Duan et al. 2013). However, shallow models require lots of expert knowledge to design and select emotion features, limiting their performance on EEG emotion classification.

Deep learning methods have been demonstrated to outperform traditional machine learning methods in many fields such as computer vision, natural language processing, and biomedical signal processing (Abbass et al. 2018; Craik et al. 2019) for the ability to learn high-level features from data automatically (Krizhevsky et al. 2012). Recently, some researchers have applied deep learning to EEG emotion recognition. Zheng et al. introduce a deep belief network (DBN) to investigate the critical frequency bands and EEG signal channels for EEG emotion recognition (Zheng and Lu 2015). Yang et al. propose a hierarchical network to classify the DE features extracted from different frequency bands (Yang et al. 2018b). Song et al. use a graph convolutional neural network to classify the DE features (Song et al. 2020). Ma et al. propose a multimodal residual Long Short-Term Memory model (MMResLSTM) for emotion recognition, which shares temporal weights across the multiple modalities (Jiaxin Ma et al. 2019). To learn the bi-hemispheric discrepancy for EEG emotion recognition, Yang et al. propose a novel bi-hemispheric discrepancy model (BiHDM) (Li et al. 2020). All those deep learning methods outperform the shallow models.

Although deep learning emotion recognition models have achieved higher accuracy than shallow models, it is still challenging to fuse more important information on different domains and capture discriminative local patterns in EEG signals. In the past decades, many researchers have investigated the critical frequency bands and channels for EEG emotion recognition. Zheng et al. demonstrate that $\beta[14\sim31\ Hz]$ and $\gamma[31\sim51\ Hz]$ bands are more related to emotion recognition than other bands, and their model achieves the best performance when combining all frequency bands. They also conduct experiments to select critical channels and propose the minimum pools of electrode sets for emotion recognition (Zheng and Lu 2015). To utilize the spatial information of EEG signals, Li et al. propose a 2D sparse map to maintain the information hidden in the electrode placement (Li et al. 2018). Zhong et al. introduce a regularized graph neural network (RGNN) to capture both local and global relations among different EEG channels for emotion recognition (Zhong et al. 2020). The temporal dependencies in EEG signals are also important to emotion recognition. For example, Ma et al. (Jiaxin Ma et al. 2019) apply LSTMs in their models to extract temporal features for emotion recognition. Shen et al. transform the DE features of different channels into 4D structures to integrate the spectral, spatial, and temporal information simultaneously and then use a four-dimensional convolutional recurrent neural network (4D-CRNN) to recognize different emotions (Shen et al. 2020). However, the differences among brain regions and frequency bands are not fully utilized in their work. To adaptively capture discriminative patterns in EEG signals, attention mechanisms have been applied to EEG emotion recognition. For instance, Tao et al. introduce a channel-wise attention mechanism, assigning the weights of different channels adaptively, along with an extended self-attention to explore the temporal dependencies of EEG signals (Tao et al. 2020). Jia et al. propose a two-stream network with attention mechanisms to adaptively focus on important patterns (Jia et al. 2020). From the above, it can be observed that it is critical to integrate information on different domains and adaptively capture important brain regions, frequency bands, and timestamps in a unified network for EEG emotion recognition.

In this paper, we propose a four-dimensional attention-based neural network named 4D-aNN for EEG emotion recognition. First, we transform raw EEG signals into 4D spatial-spectral-temporal representations which consist of several temporal slices. Different brain regions and frequency bands vary in the contributions to EEG emotion recognition, and the temporal dependencies of 4D representations should also be considered. Therefore, we employ attention mechanisms on both a CNN and a bidirectional LSTM network to adaptively capture discriminative patterns. For the CNN model, the attention mechanism is applied to the spatial and spectral dimensions of each temporal slice so that the important brain regions and frequency bands could be captured. As for the bidirectional LSTM model, the attention mechanism is applied to utilize long-range temporal dependencies so that the importance of different temporal slices in one 4D representation could be fully explored.

The primary contribution of this paper are summarized as follows: a) We propose a four-dimensional attention-based neural network, which fuses information on different domains and captures discriminative patterns in EEG signals based on the 4D spatial-spectral-temporal representation. b) We conduct experiments on the SEED dataset, and the experimental results indicate that our model achieves state-of-the-art performance under intra-subject splitting.

The remainder of this paper is organized as follows. We describe our proposed method in the *Method* section. Dataset, experiment settings, results, ablation studies, and discussion are presented in the *Experiment* section. Finally, conclusions are given in the *Conclusion* section.

## Method

Figure 1 illustrates the overall structure of 4D-aNN for EEG emotion recognition. It consists of the 4D spatial-spectral-temporal representation, the attention-based CNN, the attention-based bidirectional LSTM, and the classifier. We will describe the details of each part in sequence.

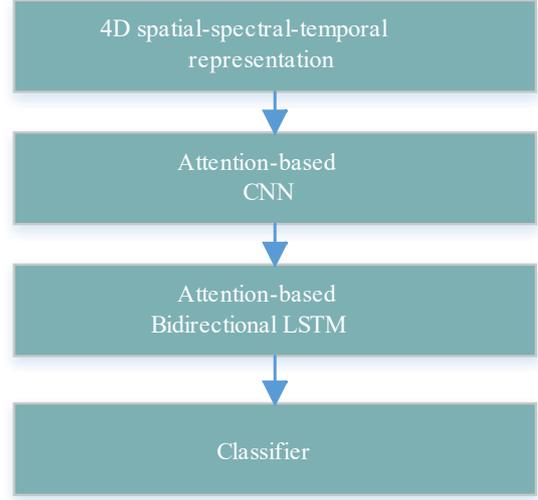

**Fig. 1** The overall structure of 4D-aNN.

## 4D spatial-spectral-temporal representation

The process of generating 4D representation is depicted in Fig. 2. As previous works do (Shen et al. 2020; Yang et al. 2018a), we split original EEG signals into $T$ seconds long segments without overlapping. Each segment is assigned with the same label as the original EEG signals. Then we decompose each segment into five frequency bands (i.e. δ[1~4 Hz], θ[4~8 Hz], α[8~14 HZ], β[14~31 Hz], and γ[31~51 Hz]) with five-order Butterworth filters. The Differential Entropy (DE) features and Power Spectral Density (PSD) features of all EEG channels, which have been proven to be effective for emotion recognition (Zheng et al. 2017), are extracted from five frequency bands respectively with a 0.5s window for each segment.

PSD is defined as

$$h_P(X) = E[x^2] \quad (1)$$

where $x$ is formally a random variable and in this context, the signal acquired from a certain frequency band on a certain EEG channel.

DE feature is capable of discriminating EEG patterns between low and high frequency energy, which is defined as

$$h_D(X) = -\int_X f(x) \log(f(x)) \, dx \quad (2)$$

where $f(x)$ is the probability density function of $x$. If $x$ obeys the Gaussian distribution $N(\mu, \sigma^2)$, DE can simply be calculated by the following formulation:

$$h_D(X) = -\int_{-\infty}^{\infty} \frac{1}{\sqrt{2\pi\sigma^2}} exp\frac{(x-\mu)^2}{2\sigma^2} \log \frac{1}{\sqrt{2\pi\sigma^2}} exp\frac{(x-\mu)^2}{2\sigma^2} dx$$
$$= \frac{1}{2}\log 2\pi e \sigma^2 \quad (3)$$

where $e$ and $\sigma$ are Euler's constant and standard deviation of $x$, respectively.

Thus, We extract a 3D feature tensor $F_n \in R^{c \times 2f \times 2T}, n = 1,2,\ldots,N$ from each segment, where $N$ is the number of total segments, $c$ is the number of EEG channels, $2f$ represents DE and PSD features of $f$ frequency bands, and $2T$ is derived by the 0.5s window without overlapping. To utilize the spatial information of electrodes, we organize all the $c$ channels as a 2D sparse map so that the 3D feature tensor $F_n$ is transformed into a 4D representation $X_n \in R^{h \times w \times 2f \times 2T}$, where $h$ and $w$ are the height and width of the 2D sparse map, respectively. The 2D sparse map of all the c channels with zero-padding is shown in Fig. 3, which preserves the topology of different electrodes. In this paper, we set $h = 19$, $w = 19$, and $f = 5$.

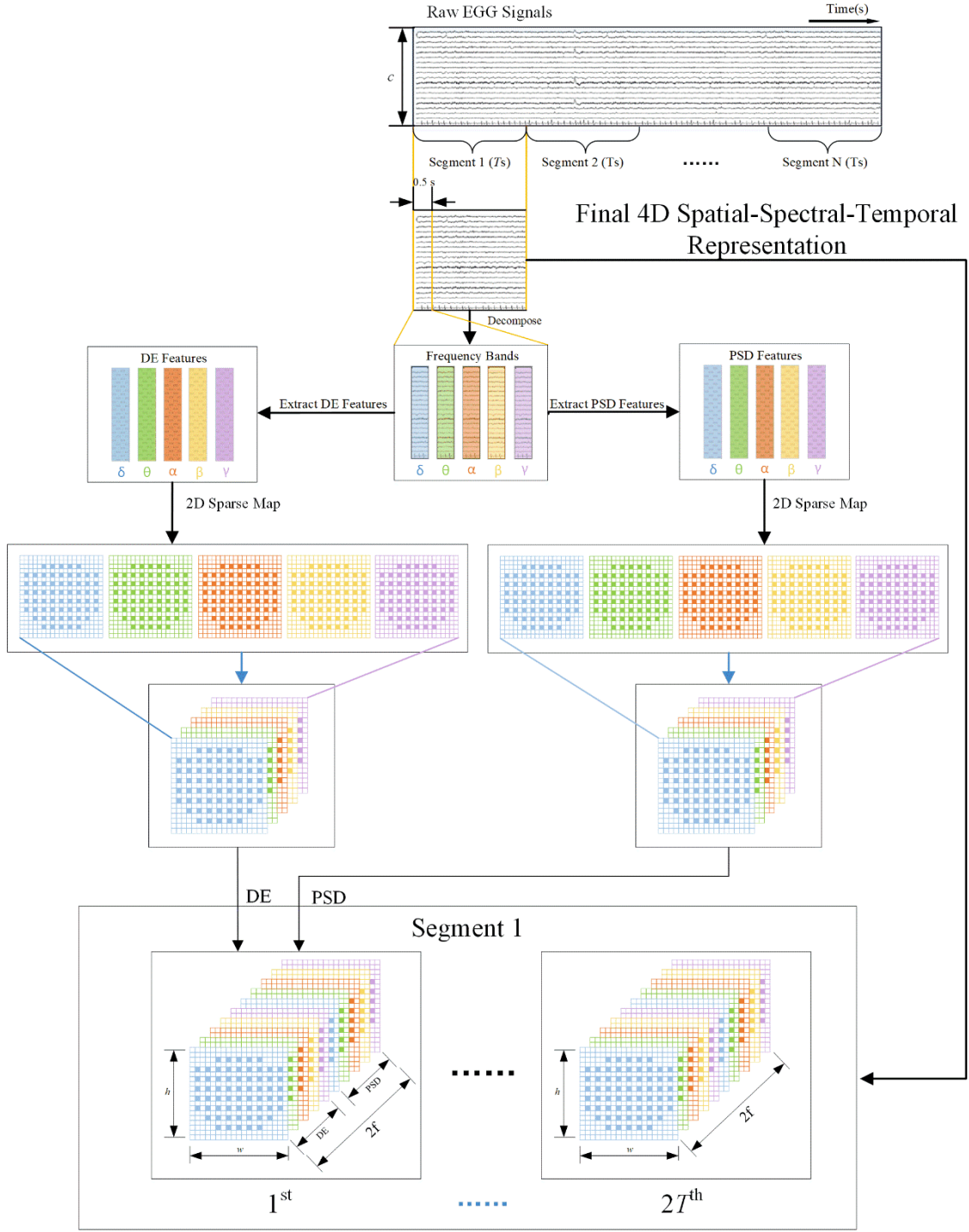

**Fig. 2** The generation of 4D spatial-spectral-temporal representation. For each $T$s EEG signal segment, we extract DE and PSD features from different channels and frequency bands with a 0.5s window. Then, the features are transformed into a 4D representation which consists of $2T$ temporal slices.

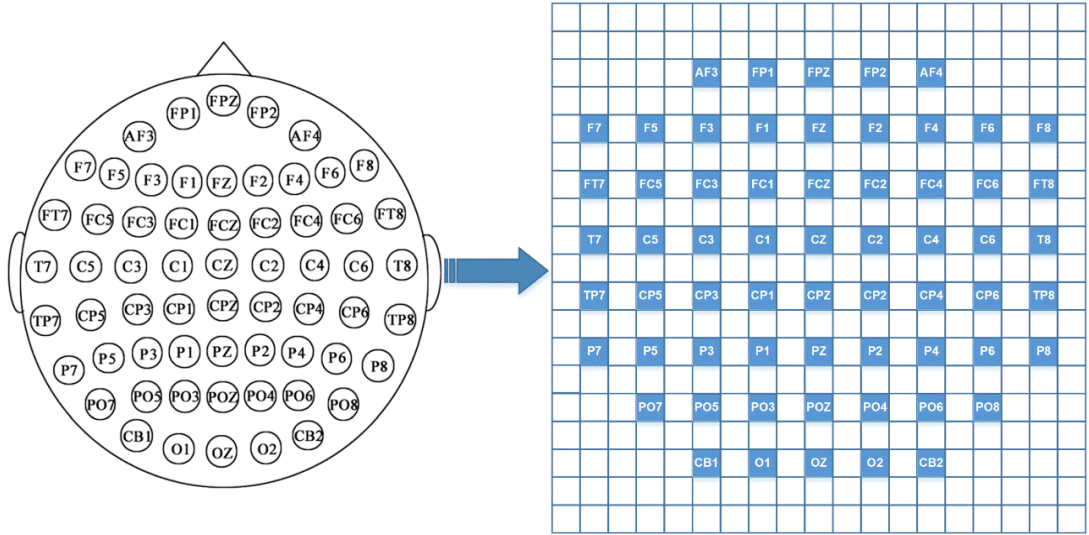

**Fig. 3** The 2D sparse map with zero-padding of 62 channels. The purpose of the organization is to preserve the positional relationships among different electrodes.

## Attention-based CNN

For a 4D spatial-spectral-temporal representation $X_n$, we extract the spatial and spectral information from each temporal slice $S_i \in R^{h \times w \times 2f}, i = 1, 2, \ldots, 2T$ with a CNN, explore the discriminative local patterns in spatial and spectral domains with a convolutional attention module, and finally get its spatial and spectral representation. The attention module here is similar to what Woo et al. propose (Woo et al. 2018), which is originally used to improve the representation power of CNN networks.

The structure of the attention-based CNN is shown in Fig. 4. It contains four convolutional layers, four convolutional attention modules, one max-pooling layer, and one fully-connected layer. The four convolutional layers have 64, 128, 256, and 64 feature maps with the filter size of $5 \times 5$, $5 \times 5$, $5 \times 5$, and $3 \times 3$, respectively. Specifically, a convolutional attention module is used after each convolutional layer to utilize the spatial and spectral attention mechanisms, and the details will be given later. We only use one max-pooling layer with a filter size of $2 \times 2$ after the last convolutional attention module to preserve more information and enhance the robustness of the network. Finally, outputs of the max-pooling layer are flattened and fed to the fully-connected layer with 150 units. Thus, for each temporal slice $S_i$, we take the final output $P_i \in R^{150}$ as its spatial and spectral representation.

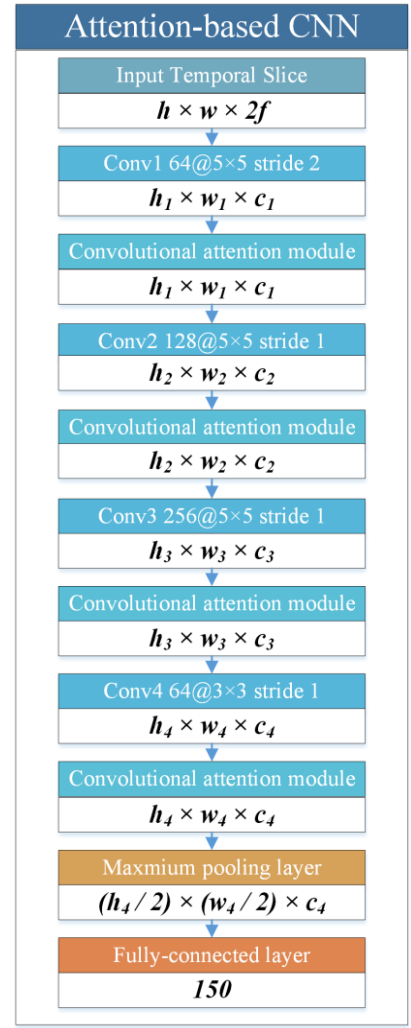

**Fig. 4** The structure of the attention-based CNN. The upper half of the blocks in the figure is the type of layers and the lower denotes the shape of its output tensors.

## Convolutional attention module

The convolutional attention module is applied after each convolutional layer to adaptively capture important brain regions and frequency bands. The structure of the convolutional attention module is shown in Fig. 5. It consists of two sub-modules, i.e. the spatial attention module and the spectral attention module.

For each convolutional layer above, its output is a 3D feature tensor $V \in R^{h_v \times w_v \times c_v}$, where $h_v$, $w_v$, and $c_v$ are the height of the 2D feature maps of $V$, the width of the 2D feature maps of $V$, and the number of the 2D feature maps of $V$, respectively. We take $V$ as the input of the convolutional attention module.

The spectral attention module is applied to identify valuable frequency bands for emotion recognition. The average pooling has been widely used to aggregate spatial information and the maximum pooling has been commonly adopted to gather distinctive features. Therefore, we shrink the spatial dimension of $V$ by a spatial-wise average pooling and a spatial-wise maximum pooling, which are defined as:

$$C_{avg,i} = \frac{1}{h_v \times w_v} \sum_{h=1}^{h_v} \sum_{w=1}^{w_v} V_i(h,w), i = 1,2,\ldots,c_v \quad (4)$$

$$C_{max,i} = max(V_i), i = 1,2,\ldots,c_v \quad (5)$$

where $V_i \in R^{h_v \times w_v}$ denotes the 2D feature map in the *i-th* channel of $V$, $C_{avg,i}$ represents the element in the *i-th* channel of the spatial average representation $C_{avg} \in R^{c_v}$, $max(Z)$ returns the largest element in $Z$, and $C_{max,i}$ is the element in the *i-th* channel of the spatial maximum representation $C_{max} \in R^{c_v}$. Subsequently, we implement the spectral attention by two fully-connected layers, a *Relu* activation function and a *sigmoid* activation function, which is defined as:

$$A_{spectral,avg} = W_2^S(Relu(W_1^S C_{avg})) \quad (6)$$
$$A_{spectral,max} = W_2^S(Relu(W_1^S C_{max})) \quad (7)$$
$$A_{spectral} = sigmoid(A_{spectral,avg} \oplus A_{spectral,max}) \quad (8)$$
$$Relu(x) = max(x,0) \quad (9)$$
$$sigmoid(x) = \frac{1}{1 + e^{-x}} \quad (10)$$

where $W_1^S$ and $W_2^S$ are learnable parameters, $\oplus$ denotes the element-wise addition, and $A_{spectral} \in R^{1 \times 1 \times c_v}$ is the spectral attention. The elements of $A_{spectral}$ represent the importance of the corresponding 2D feature maps of the spectral domain. After generating the spectral attention $A_{spectral}$, the output of the spectral attention module can be defined as:

$$V' = A_{spectral} \otimes V \quad (11)$$

where $V'$ denotes the refined 3D feature tensor, and $\otimes$ represents the element-wise multiplication.

The spatial attention module is applied to identify valuable brain regions for emotion recognition. Firstly, we shrink the spectral dimension of $V'$ by spectral-wise average pooling and spectral-wise maximum pooling, which is defined as:

$$SPA_{avg,(h,w)} = \frac{1}{c_v} \sum_{c=1}^{c_v} S'_{h,w}(c), h = 1,2,\ldots,h_v; w = 1,2,\ldots,w_v \quad (12)$$

$$SPA_{max,(h,w)} = max(S'_{h,w}), h = 1,2,\ldots,h_v; w = 1,2,\ldots,w_v \quad (13)$$

where $S'_{h,w} \in R^{c_v}$ denotes the channel in the *h-th* row and *w-th* column of $V'$, $SPA_{avg,(h,w)}$ represents the element in the *h-th* row and *w-th* column of the spectral average representation $SPA_{avg} \in R^{h_v \times w_v \times 1}$ and $SPA_{max,(h,w)}$ is the element in the *h-th* row and *w-th* column of the spectral maximum representation $SPA_{max} \in R^{h_v \times w_v \times 1}$. In the following, we implement the spatial attention with a convolutional layer and a *sigmoid* activation function, which is defined as:

$$SPA = Cat(SPA_{avg}, SPA_{max}) \quad (14)$$
$$A_{spatial} = Sigmoid(Conv(SPA)) \quad (15)$$

where $Cat(SPA_{avg}, SPA_{max})$ denotes the concatenation of $SPA_{avg}$ and $SPA_{max}$ along the spectral dimension, $Conv(SPA)$ represents the convolutional layer for $SPA$, and $A_{spatial} \in R^{h_v \times w_v \times 1}$ is the spatial attention. The elements of $A_{spatial}$ represent the importance of the corresponding regions of the spatial domain. Subsequently, the output of the spatial attention module can be defined as:

$$V'' = A_{spatial} \otimes V' \quad (16)$$

where $V'' \in R^{h_v \times w_v \times c_v}$ denotes the final output 3D feature tensor of the convolutional attention module.

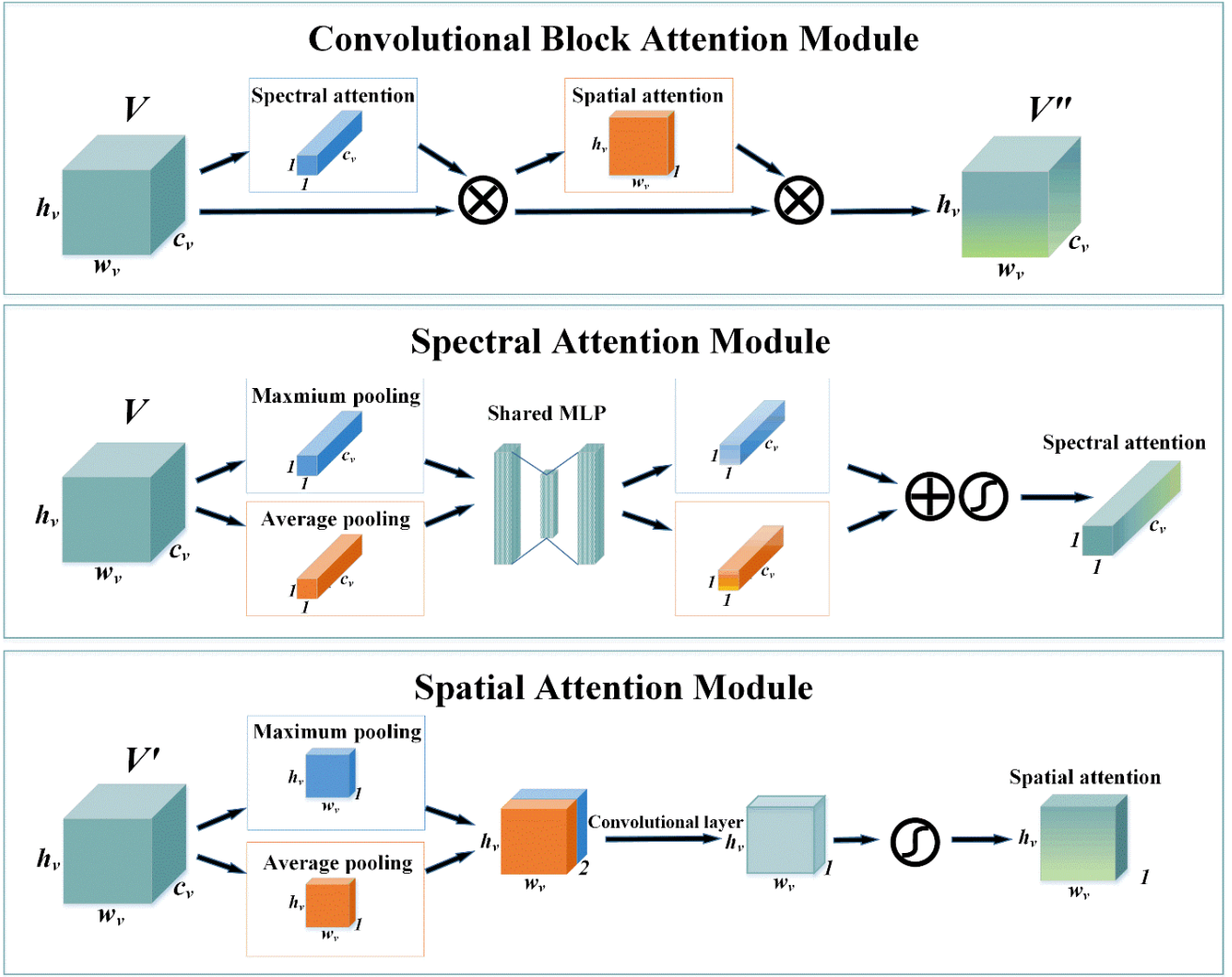

**Fig. 5** The top block is the overall structure of the convolutional attention block, it consists of the spectral attention module and the spatial attention module. The middle block represents the generation of spectral attention. The bottom block denotes the generation of spatial attention.

**Attention-based bidirectional LSTM**

For each temporal slice $S_i \in R^{h \times w \times 2f}, i = 1, 2, \ldots, 2T$, the final output of the attention-based CNN is $P_i \in R^{150}$. Since the variation between different temporal slices contains temporal information for emotion recognition, we utilize an attention-based bidirectional LSTM to explore the importance of different slices, as shown in Fig. 6.

A bidirectional LSTM connects two unidirectional LSTMs with opposite directions to the same output. Comparing with a unidirectional LSTM, a bidirectional LSTM preserves information from both past and future, making it understand the context better. In this paper, the bidirectional LSTM comprises two unidirectional LSTMs with 36 memory cells. The unidirectional LSTM for positive time direction, LSTM$_P$ takes the output sequence of the attention-based CNN $P^P = (P_1, P_2, \ldots, P_{2T})$ as the input sequence, while the other for negative time direction, LSTM$_N$ takes the reverse sequence $P^N = (P_{2T}, P_{2T-1}, \ldots, P_1)$ as the input sequence. The outputs of the $i$-th node of the unidirectional LSTMs are $Y_i^P \in R^{36}$ and $Y_i^N \in R^{36}, i = 1, 2, \ldots, 2T$, respectively. Then, we concatenate $Y_i^P$ and $Y_{2T+1-i}^N$ as the output of the $i$-th node of the bidirectional LSTM $Y_i \in R^{72}$. Different from traditional ways that only use the output of the last node of an LSTM for classification or other applications, we take the outputs of all the bidirectional LSTM nodes $Y \in R^{2T \times 72}$ into consideration and explore the importance of different temporal slices by the temporal attention mechanism.

The temporal attention mechanism is implemented with two fully-connected layers, a *Relu* activation function, and a *softmax* activation function, which is defined as:

$$Tem_i = W_2^T\left(Relu(W_1^T Y_i + b_1^T)\right) + b_2^T \quad (17)$$

$$A_{temporal} = softmax(Tem) \quad (18)$$

$$softmax(x) = \frac{exp(x)}{\sum exp(x)} \quad (19)$$

where $W_1^T$, $W_2^T$, $b_1^T$, and $b_2^T$ are learnable parameters, $Tem_i$ represents the *i-th* element of $Tem \in R^{2T \times 1}$ which projects $Y \in R^{2T \times 72}$ to a lower dimension, and $A_{temporal} \in R^{2T \times 1}$ is the temporal attention. The elements of $A_{temporal}$ represent the importance of the corresponding temporal slices. Subsequently, the high-level representation of the 4D sample $X_n$ can be defined as:

$$L_n(e) = \sum A_{temporal} \otimes Y_e, e = 1, 2, \ldots, 72 \quad (20)$$

where $Y_e \in R^{2T \times 1}$ denotes the *e-th* column of $Y \in R^{2T \times 72}$ and $L_n(e)$ is the *e-th* element of the high-level representation $L_n \in R^{72}$, which integrates spatial, spectral, and temporal information of $X_n$.

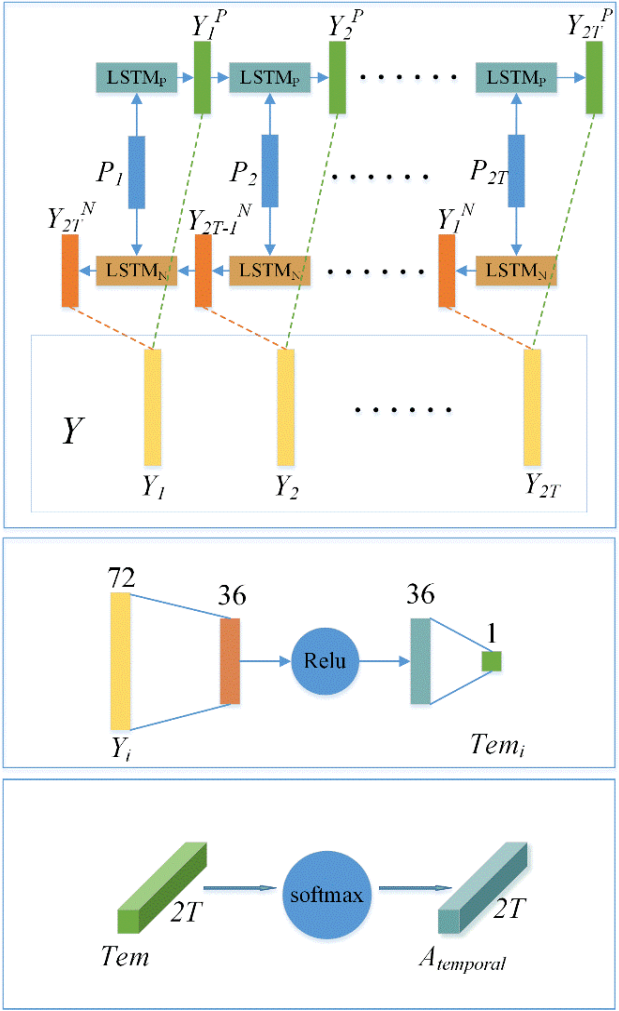

**Fig. 6** The top block is the structure of the bidirectional LSTM. We concatenate the outputs of LSTM_P and LSTM_P as the output of the bidirectional LSTM, $Y \in R^{2T \times 72}$. The middle block represents the projection of the outputs of the bidirectional LSTM. The bottom block denotes the generation of temporal attention.

### Classifier

Based on the high-level representation $L_n$ of EEG signals, we apply a fully-connected layer and a *softmax* activation function to predict the label of the 4D sample $X_n$, which can be defined as follows:

$$Pre = softmax(W^p L_n + b^p) \quad (21)$$

where $W^p, b^p$ are learnable parameters and $Pre \in R^C$ denotes the probability of $X_n$ belonging to all the $C$ classes. Specifically, the class of the largest probability is the predicted label of 4D-aNN.

## Experiment

In this section, we firstly introduce a widely used dataset. Then, the experiment settings are described. Finally, the results on the dataset are reported and discussed.

### SEED Dataset

SEED dataset (Zheng and Lu 2015) contains 3 different categories of emotion data: positive, neutral, and negative. For each kind of emotion, 5 film clips that are about 4 minutes long and can elicit the desired target emotion are selected. 15 healthy subjects (7 males and 8 females, with age (23.27 ± 2.37)) take part in the EEG signals collection experiment. 3 groups of experiments are conducted for each subject, and each experiment consists of 15 clips viewing processes. Each clip viewing process can be divided into four stages, including a 5 seconds hint of start, a 4 minutes clip period, a 45 seconds self-assessment, and a 15 seconds rest period. The order of the 15 clips is arranged so that two clips eliciting the same emotion are not shown consecutively. The EEG signals in the experiments are recorded by a 62-channel's ESI NeuroScan system and down-sampled to 200 Hz. Besides, the EEG signals seriously contaminated by electromyography (EMG) and electrooculography (EOG) are removed manually. Then, a bandpass filter between 0.3 to 50 Hz is applied to filter the noise.

### Settings

The proposed 4D-aNN takes a 4D segment $X_n \in R^{h \times w \times 2f \times 2T}$ as the input. In this paper, we adopt the 2D sparse map with $h = 19$ and $w = 19$ to maintain the positional relationship of electrodes. As shown in previous works, the combination of all the 5 bands can contribute to better results so that we set $f = 5$. For each experiment, we set the length of segments $T$ as 3, obtaining about 1128 samples per experiment. Then, we conduct a fivefold cross-validation on each experiment and calculate the average classification accuracy (ACC) and standard deviation (STD) of 3 experiments for each subject. The average ACC and STD of all subjects are taken as the final performances of our method. We train the 4D-aNN on an NVIDIA GTX 1080 GPU. The Adam optimization is applied to minimize the loss function. We set the learning rate as 0.0003, the batch size as 12, and the maximum of epochs as 150.

**Baseline Models**

- HCNN (Li et al. 2018): It uses a hierarchical CNN architecture for EEG emotion recognition, taking 2D DE feature maps extracted from γ band as inputs. HCNN only considers the spatial information of EEG signals.
- BiHDM (Li et al. 2020): It considers the asymmetric differences between two hemispheres for EEG emotion recognition.
- RGNN (Zhong et al. 2020): It takes the biological topology among different brain regions into consideration to capture both global and local relations among different EEG channels.
- 4D-CRNN (Shen et al. 2020): It builds DE features extracted from EEG signals into 4D feature structures and uses a convolutional recurrent neural network to extract spatial features, spectral features, and temporal features for EEG emotion recognition.
- SST-EmotionNet (Jia et al. 2020): It uses a two-stream network to extract spatial, spectral, and temporal features. Besides, SST-EmotionNet adopts the attention mechanisms to improve its performance on EEG emotion recognition.

**Results**

We compare our model with 5 baseline models on SEED dataset. Table 1 presents the average ACC and STD of these models for EEG emotion recognition. HCNN uses the hierarchical CNN architecture to classify emotion, but only considers the spatial information of EEG signals, reaching 88.60% on classification accuracy. BiHDM (Li et al. 2020) applies four directed RNNs to obtain the deep representation of all the EEG electrodes' signals, reaching 93.12% on classification accuracy. RGNN considers the biological topology among different brain regions, reaching 94.24% on classification accuracy. 4D-CRNN takes 4D DE feature maps containing spatial, spectral, and temporal information as inputs, reaching 94.74% on classification accuracy. SST-EmotionNet uses a two-stream network with the attention mechanisms, reaching 96.02% on classification accuracy. However, the data size of each input sample of SST-EmotionNet is about 4 times larger than 4D-aNN. Comparing with the baseline models, the proposed 4D-aNN achieves the state-of-the-art performance on the SEED dataset under intra-subject splitting. The average ACC of all subjects is 96.10%. The performances on each subject are shown as Fig. 7, and there are 9 subjects (#5, #6, #8, #9, #10, #11, #12, #13, and #15) whose performances are better than the average ACC. Specifically, to make a fair comparison with 4D-CRNN, we conduct experiments on 4D-aNN (DE) and 4D-aNN (PSD), which represents the 4D-aNN only takes DE features as inputs and only takes PSD features as inputs, respectively. The accuracy of 4D-aNN (DE) exceeds that of 4D-CRNN by 0.65%, indicating the superiority of the proposed 4D-aNN. When compared with 4D-aNN (DE) and 4D-aNN (PSD), 4D-aNN displays the best performance, which indicates the effectiveness of the combination of different features.

**Table 1** The performance (average ACC and STD (%)) of the compared models.

| Model | SEED | |
|---|---|---|
| | ACC (%) | STD (%) |
| HCNN (Li et al. 2018) | 88.60 | 2.60 |
| BiHDM (Li et al. 2020) | 93.12 | 6.06 |
| RGNN (Zhong et al. 2020) | 94.24 | 5.95 |
| 4D-CRNN (Shen et al. 2020) | 94.74 | 2.32 |
| SST-EmotionNet (Jia et al. 2020) | 96.02 | 2.17 |
| 4D-aNN | 96.10 | 2.61 |
| 4D-aNN (DE) | 95.39 | 3.05 |
| 4D-aNN (PSD) | 90.49 | 7.97 |

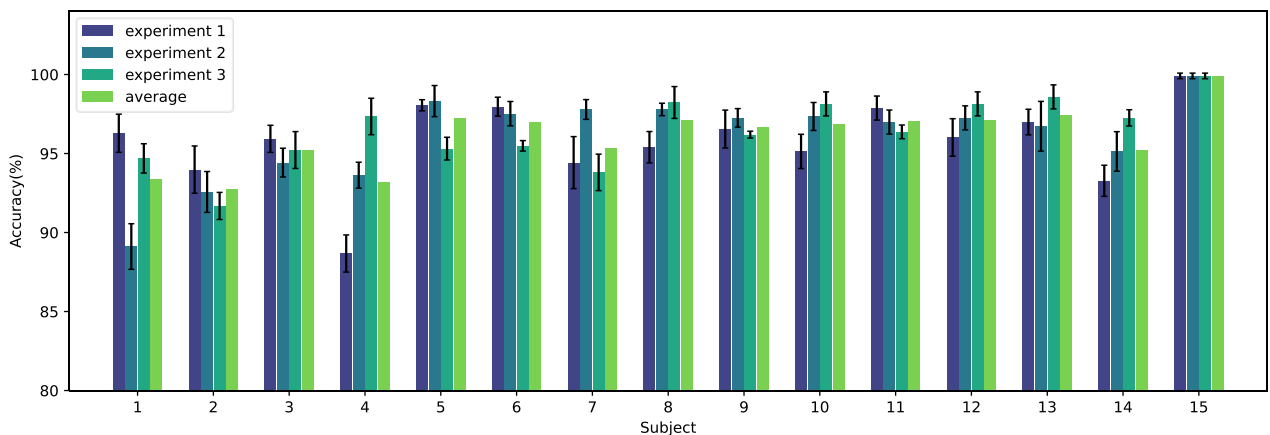

**Fig. 7** The performance of 4D-aNN on each subject. In the SEED dataset, 3 experiments are conducted for each subject. We evaluate the performance of each experiment and also present the average classification accuracy for each subject.

To verify the importance of the attention mechanisms in our model, we conduct an additional experiment for ablation studies on SEED dataset. The experiment is ablation on spatial, spectral, and temporal attention mechanisms. We evaluate the performances of 4D-aNN when spatial, spectral, temporal, and all the attention mechanisms are ablated respectively. As shown in Fig. 8, when one of the attention mechanisms is ablated, the classification accuracy decreases. 4D-aNN without the spectral attention mechanism decreases by 0.63%, 4D-aNN without the spatial attention mechanism decreases by 0.47%, and 4D-aNN without the temporal attention mechanism decrease by 1.19%. Specifically, 4D-aNN without all the attention mechanisms decreases by 2.17%, which is the worst among the models used for comparison. In conclusion, the results indicate that the attention mechanisms make contributions to EEG emotion recognition for the ability to capture the discriminative local patterns in spatial, spectral, and temporal domains.

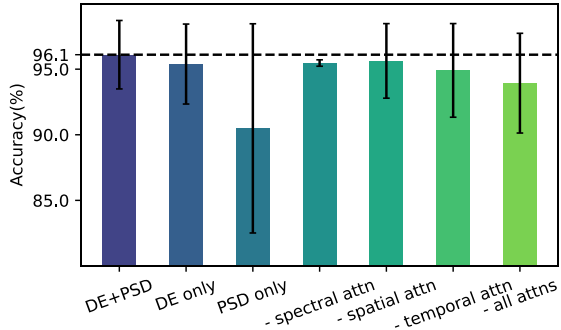

**Fig. 8** Ablation studies on different input features and attention modules of 4D-aNN. "–" denotes the ablation on certain attention modules.

In particular, to explore the critical brain regions for different emotions, we separately depict the electrode activity heatmaps in Fig. 9. We draw the heatmaps using *Grad-CAM++* (Chattopadhay et al. 2018), based on the experimental results of subject #15. *Grad-CAM++* uses the last convolutional layer feature maps and the class scores of the classifier to generate heatmaps. The heatmaps are able to explain which input regions are important for predictions. In this work, the size of each heatmap is 19×19, which is the same as the 2D sparse map. The elements in the heatmaps represent the contributions of the corresponding brain regions to the recognition of the target emotions. From Fig. 9, We can observe the distinct distributions of important brain regions with regard to different emotions: channels *FC5*, *FC3*, and *C5* are important for recognition of positive emotions, channels *CP5*, *CP3*, and *CP1* are important for recognition of neutral emotions, and channels *PO7*, *PO5*, and *P3* are important for recognition of negative emotions for subject #15. In particular, the critical brain regions could vary with different subjects, time, and emotions so that the attention mechanisms that enable 4D-aNN to adaptively capture discriminative patterns make sense for EEG emotion recognition.

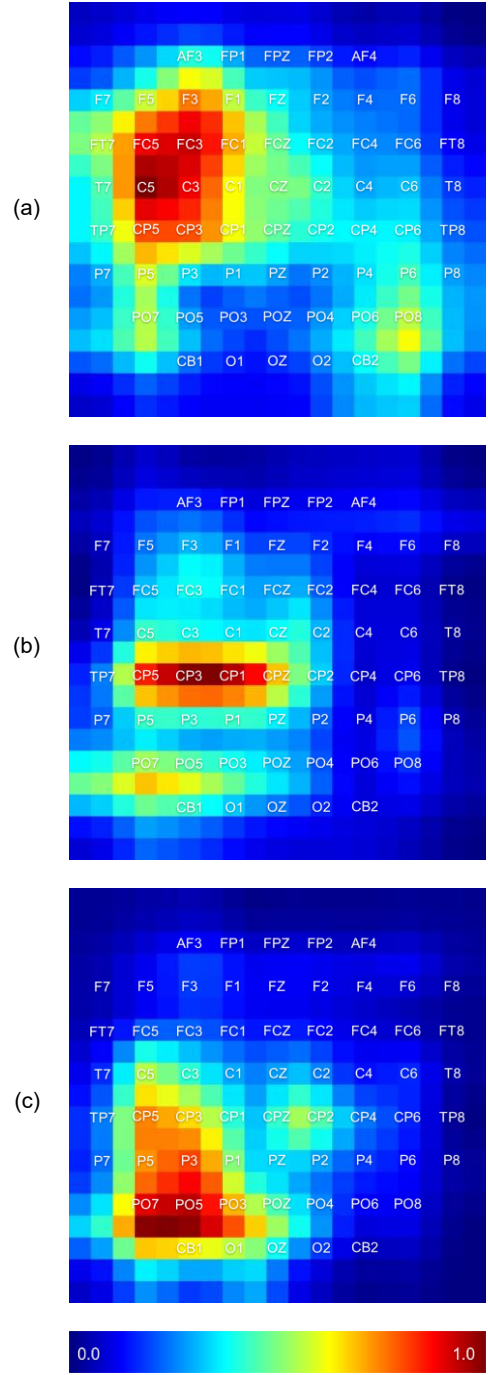

**Fig. 9** The electrode activity heatmaps based on the experimental results of subject #15. Parts (a), (b), and (c) correspond to positive, neutral, and negative emotions, respectively. Dark red regions denote more significant contributions to the recognition of the corresponding emotions.

## Discussion

We conduct several experiments to investigate the use of 4D-aNN which fuses the spatial-spectral-temporal information and the effectiveness of the attention mechanisms on different domains for EEG emotion classification. In this section, we discuss three noteworthy points.

First, to deal with the spatial-spectral information, we apply an attention-based CNN which consists of a CNN network, a spectral attention module, and a spatial attention module. The CNN network extracts the spatial-spectral representation from inputs first. Then, the spectral attention mechanism is applied to each spectral feature to explore the importance of different frequency bands and features. Besides, the spatial attention mechanism is applied to each 2D feature map to adaptively capture the critical brain regions. The critical brain regions and frequency bands could vary with different individuals, emotions, and time so that the ability to capture discriminative patterns of the attention modules improves the performance of 4D-aNN.

Second, to explore the temporal dependencies in 4D spatial-spectral-temporal representations, we utilize an attention-based bidirectional LSTM. The bidirectional LSTM extracts high-level representations from the outputs of the attention-based CNN. Different from traditional ways that only use the output of the last node of an LSTM for classifications or other applications, we consider outputs of all the nodes with the temporal attention mechanism. The temporal attention mechanism adaptively assigns weights of different temporal slices so that the dynamic content of emotions in 4D representations could be captured better.

Third, to address the importance of the attention mechanisms, we conduct ablation studies on different attention modules. 4D-aNN without the spatial, spectral, and temporal attention mechanism decreases by 0.47%, 0.63%, and 1.19% on classification accuracy, respectively. In particular, 4D-aNN without all the attention mechanisms decreases by 2.17%, which is the worst among the models in comparison. The experimental results demonstrate the effectiveness of the attention mechanisms to adaptively capture discriminative patterns.

## Conclusion

In this paper, we propose the 4D-aNN model for EEG emotion recognition. The 4D-aNN takes 4D spatial-spectral-temporal representations containing spatial, spectral, and temporal information of EEG signals as inputs. We integrate the attention mechanisms into the CNN module and the bidirectional LSTM module. The CNN module deals with the spatial and spectral information of EEG signals while the spatial and spectral attention mechanisms capture critical brain regions and frequency bands adaptively. The bidirectional LSTM module extracts temporal dependencies on the outputs of the CNN module while the temporal attention mechanism explores the importance of different temporal slices. The experiments on SEED dataset demonstrate better performance than all baselines. In particular, the ablation studies on different attention modules show the effectiveness of the attention mechanisms in our model for EEG emotion recognition.